\documentclass[11pt,a4paper]{article}
\usepackage[hyperref]{acl2021}
\usepackage{times}
\usepackage{graphicx}
\usepackage{latexsym}
\usepackage{float}
\usepackage{footnote}
\usepackage{multirow}
\usepackage{array}

\usepackage{microtype}

\aclfinalcopy 
\def\aclpaperid{1627} 

\title{Multi-stage Pre-training over Simplified Multimodal Pre-training Models}

\author{Tongtong Liu \\
  Beijing University of Posts \\
  and Telecommunications \\
  \texttt{ttliu@bupt.edu.cn} \\\And
  Fangxiang Feng \\
  Beijing University of Posts \\
  and Telecommunications \\
  \texttt{fxfeng@bupt.edu.cn} \\\And
  Xiaojie Wang \\
  Beijing University of Posts \\
  and Telecommunications \\
  \texttt{xjwang@bupt.edu.cn} \\
  }
\date{}

\begin{document}
\maketitle
\begin{abstract}
Multimodal pre-training models, such as LXMERT, have achieved excellent results in downstream tasks. However, current pre-trained models require large amounts of training data and have huge model sizes, which make them difficult to apply in low-resource situations. How to obtain similar or even better performance than a larger model under the premise of less pre-training data and smaller model size has become an important problem. In this paper, we propose a new \textbf{M}ulti-\textbf{s}tage \textbf{P}re-training (MSP) method, which uses information at different granularities from word, phrase to sentence in both texts and images to pre-train the model in stages. We also design several different pre-training tasks suitable for the information granularity in different stage in order to efficiently capture the diverse knowledge from a limited corpus. We take a Simplified LXMERT (LXMERT-S), which has only 45.9\% parameters of the original LXMERT model and 11.76\% of the original pre-training data as the testbed of our MSP method. Experimental results show that our method achieves comparable performance to the original LXMERT model in all downstream tasks, and even outperforms the original model in Image-Text Retrieval task.
\end{abstract}
\section{Introduction}
Self-attention based Transformer ~\citep{Vaswani:17} effectively overcomes the problem of RNN being difficult to run in parallel, and greatly promotes the development of large-scale pre-training models. The pre-training language models, such as BERT ~\citep{Devlin:18}, have achieved excellent performance in many natural language processing tasks. With their big success, researchers have also developed pre-training models on multimodal tasks. A series of multimodal pre-training models have been proposed, such as ViLBERT~\citep{Batra:19}, LXMERT~\citep{Tan:19}, UNITER~\citep{Chen:19} etc., and have achieved excellent results in language-vision multimodal tasks.

However, the current pre-training models are normally with large-scale parameters, require huge pre-training data and have very high demands on computational resources. For example, the GPT model~\citep{radford:18} has 110 Million parameters, GPT-2~\citep{radford:19} has 1.5 Billion parameters, and GPT-3~\citep{brown:20} has a staggering 175 Billion parameters. The same is true for multimodal pre-trained models. For example, LXMERT~\citep{Tan:19} has 183.5 Million parameters and requires 816 TitanX GPU hours for training on 9.18 Million text-image pairs. The sizes of these models are too huge for them to be deployed in many real-world scenarios. Therefore, the study of lightweight pre-training models, which can achieve similar performances to large-scale models with smaller parameter scales and training costs, is significantly valuable.

There are some types of work on developing lightweight pre-trained models, including the design of the model structure, quantization, pruning and distillation. For example, ALBERT~\citep{Chi:20} is a lightweight model through structural design such as parameter sharing and parameter decomposition, and achieves better performance than original models; Q8BERT~\citep{Zafrir:19} compresses the model to 1/4 of the original model but with no more than 1\% performance loss by quantizing 32bit floating point into 8bit; ~\citep{Michel:19} used BERT weight pruning to compress the model and found that removing a large number of attention heads would not have a major impact on the model performance; TinyBERT~\citep{Jiao:19} reduced the model size by 7.5 times but with no more than 4\% performance loss by designing a teacher-student distillation model.

All above works are on language pre-training models, and most of them concern scales of model parameters. There are few works on cutting training data and light weighing multimodal pre-training model. In fact, compared with language model, multimodal pre-training models should deal with data from both language and visual modal, which demand larger amounts of data and more computational resources. Meanwhile, collections of training data are more difficult. Taking for example the size of text-image pairs used for multimodal pre-training, the frequently used MS COCO~\citep{lin:14} is a high quality dataset with only 0.82M pairs, while LAIT~\citep{qi:20} is already a big data with 10M pairs but with average quality. Therefore, it is significantly valuable to develop lightweight multimodal pre-training models which can make use of limited data efficiently.

Existing research on curriculum learning~\citep{bengio:09} has shown that imitating the process of human learning by gradually increasing the difficulty of a task from simple to complex in stages helps to make better use of different types of data and effectively improve the performance of learning. Many models~\citep{qi:20} use as much as data available but few works have been done on how to arrange the tasks for better making use of limited data. We therefore borrow the idea of curriculum learning on training pre-training models. We construct a pre-training process which makes use of data from smaller units to bigger units in stages, and design appropriate pre-training tasks for each corresponding stage. 
\begin{figure*} 
\centering 
\includegraphics[width=1.0\textwidth]{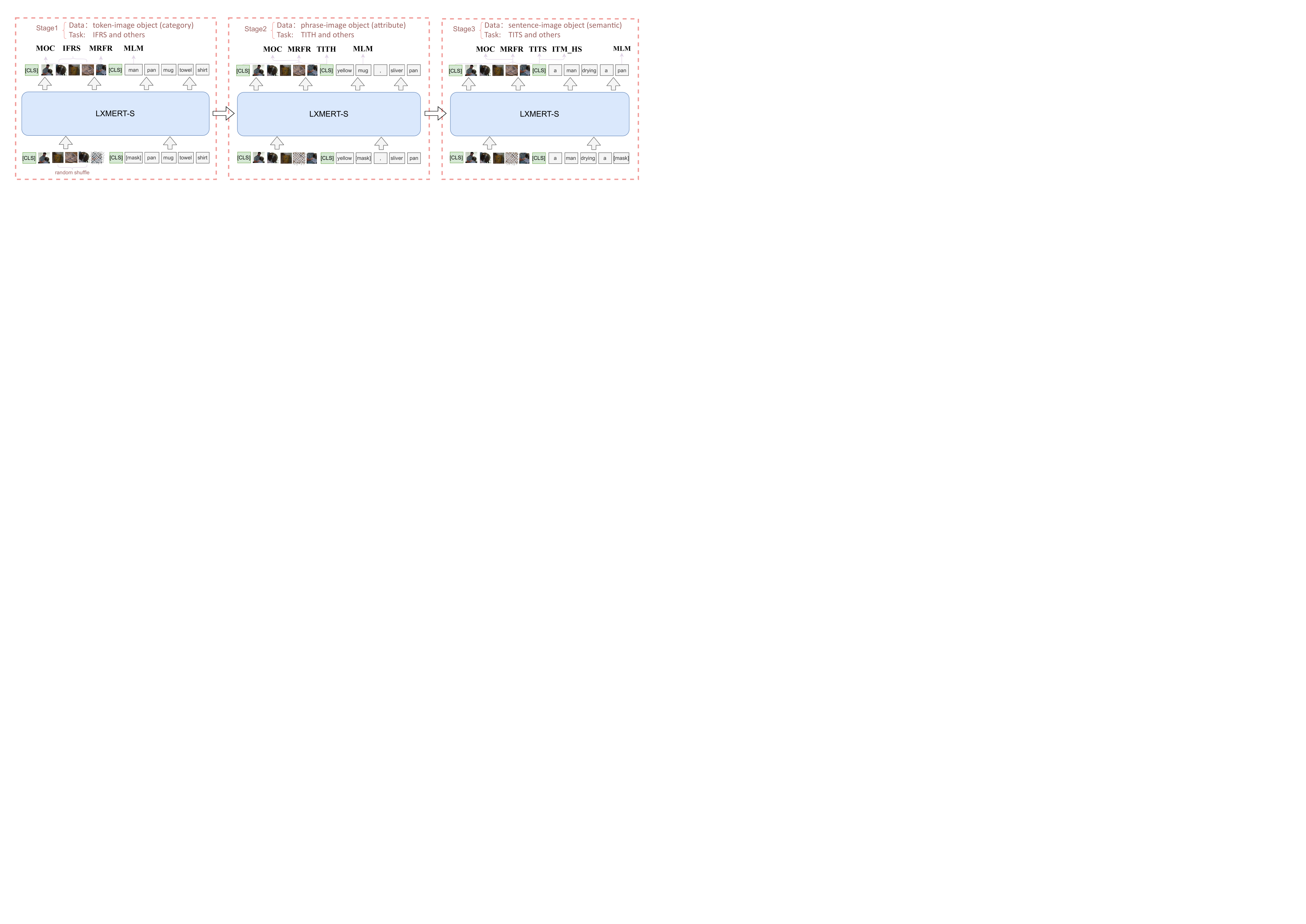} 
\caption{Overview of our proposed MSP method, including three stages from token, phrase to sentence-based pre-training, with appropriate pre-training tasks for each stage of pre-training.} 
\label{Fig.main1} 
\end{figure*}

Specifically, we propose a new \textbf{M}ulti-\textbf{s}tage \textbf{P}re-training (MSP) method. The first pre-training stage is on the token units, where the text input is the category labels of the objects in the images, and the image input is the object features. An Image Features Random Shuffle (IFRS) is designed as a pre-training task for this stage. IFRS randomly shuffles the object features, and the model predicts the original object order based on the text information. The second stage focuses on phrase units. Phrase-level descriptions of the image are input on the text side and image features are input on the image side. A Topic of Image and Text for Phrase (TITP) task is designed for it. The third stage is sentence-based pre-training. Sentence-level captions are input on the text side, and image features are input on the image side. A Topic of Image and Text for Sentence (TITS) task is designed for it. We take a Simplified LXMERT (LXMERT-S) which has fewer parameters and less pre-training data as the testbed of our MSP method. Experimental results show that our method achieves comparable performance to the original LXMERT model in downstream tasks.

The main contributions of our work are as follows: (1) We propose a new MSP method that allows the model to learn different granularities of text-image correspondence information at different stages; (2) For each stage, we design pre-training tasks suitable for that stage, IFRS task for token-based pre-training, TITP task for phrase-based pre-training, and TITS task for sentence-based pre-training; (3) With less pre-trained data (11.76\%), fewer model parameters (45.9\%), less resource consumption (25\%) and less training time (46.57\%), the performances of downstream tasks are comparable to or even exceed that of the original model.

\section{Related Works}

\paragraph{Multimodal Pre-training Models} Multimodal pre-training models are mainly divided into two categories: single-stream models and two-stream models. Single-stream models such as  B2T2~\citep{alberti:19}, OSCAR~\citep{li:20}, etc., fuse image and text information at the beginning of the input; two-stream models such as ViLBERT~\citep{Batra:19}, LXMERT\citep{Tan:19}, etc., encode the image and text information alone first and then fuse them later. Generally two-stream models will have more parameters than single-stream models, but whether the single-stream model or the two-stream model has better performance or is related to the specific tasks require more rigorous experimental proof. We conduct follow-up experiments based on the two-stream model LXMERT by removing the coding layer of the individual modalities and keeping only the fusion coding layer, so that the simplified LXMERT model is more like the single-stream model.
\paragraph{Multimodal Pre-training Data}There are several different considerations on making use of data. VisualBERT~\citep{li:19} believes that pre-training on the target dataset can improve the performance of the model, so VisualBERT first pre-trains on COCO Caption and then continues pre-training on the target dataset (e.g. VQA). ImageBERT~\citep{qi:20}, on the other hand, is trained on the out-of-domain LAIT dataset and then on the in-domain datasets, such as Conceptual Captions(CC)~\citep{Sharma:18} and SBU Captions~\citep{Ordonez:11}. It can be said the dataset that is most similar to the downstream task is used for training at last, and the general data is used firstly. Clearly, this way of using data is directly related to the downstream tasks. Different downstream tasks might lead to different order of data usage. In this paper, we design a staged pre-training from word-level to phrase-level to sentence-level, which is related to the size of information units. We also design suitable pre-training tasks for different phases to fully exploit the text-image information correspondence of different units in each phase, which has consistent effectiveness for different downstream tasks.
\paragraph{Multimodal Pre-training Tasks}The mostly employed language pre-training task is Masked Language Modeling (MLM)~\citep{Chen:19}, where tokens are masked with a probability and those masked tokens are predicted by the model. Masked Region Feature Regression (MRFR)~\citep{Chen:19}, which is similar to the MLM task, is a popular image pre-training task. Masked Object Classification (MOC)~\citep{qi:20} task can be regarded as a multimodal pre-training task, which is to predict the category label of each masked object feature. Another popular multimodal pre-training task called Image-Text Matching (ITM)~\citep{Chen:19} is similar to the Next Sentence Prediction (NSP) task in BERT~\citep{Devlin:18}, where an image corresponding to a text is randomly replaced with a probability of 50\%, and the task is to discriminate whether the image matches the text. The existing pre-training tasks for multimodal data are limited. We design new pre-training tasks with the aim of making full use of the existing training dataset at different granularities.
\section{Method}
The overall structure of our MSP method is shown in Figure \ref{Fig.main1}. The pre-training process is divided into three stages based on different granularities of text-image correspondence from token, phrase to sentence. We design corresponding pre-training tasks for the three stages.

\begin{figure} 
\centering 
\includegraphics[width=0.45\textwidth]{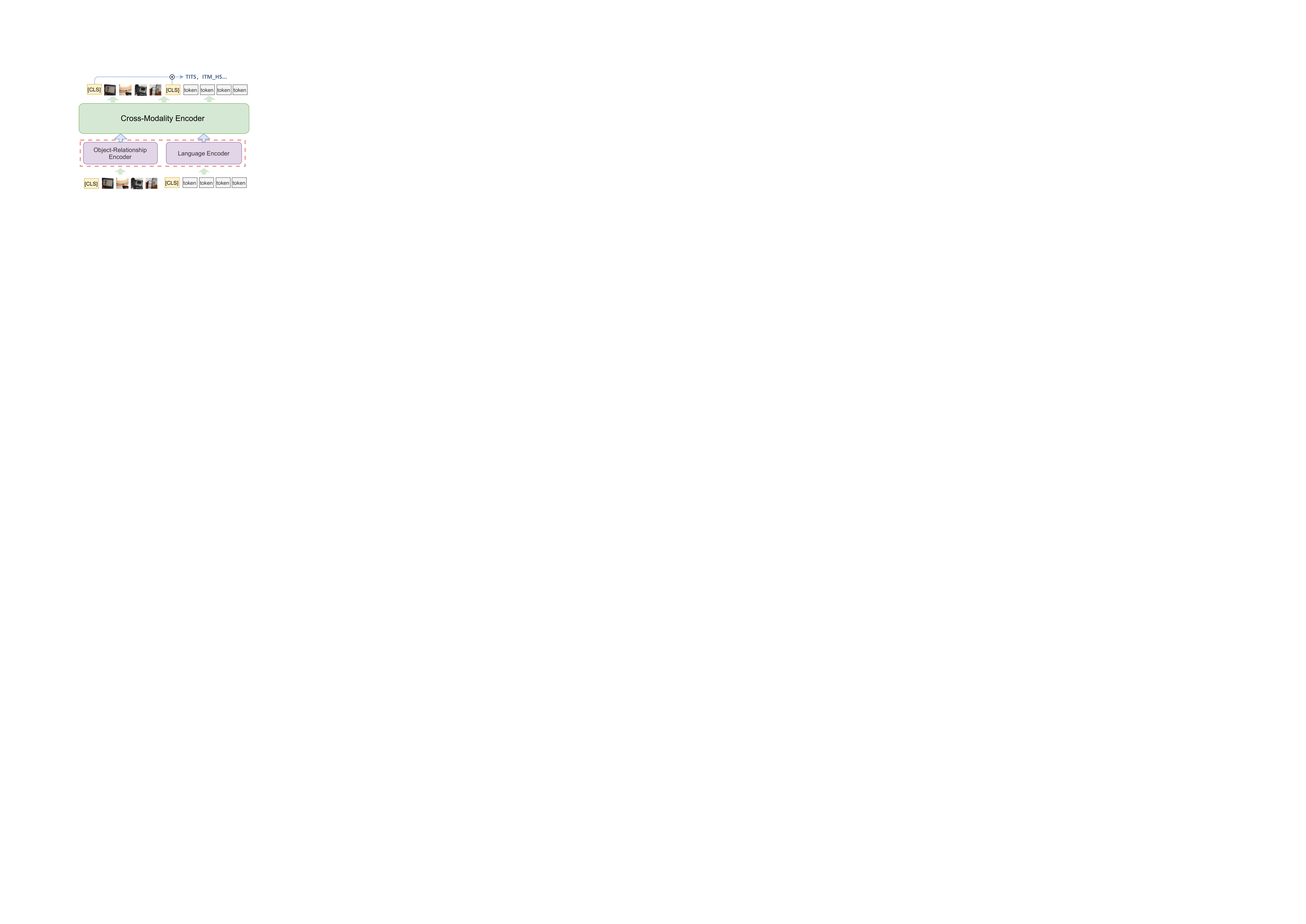} 
\caption{Overview of the simplified process of LXMERT. We obtained a Simplified LXMERT (LXMERT-S) by removing the Object-Relationship Encoder and Language Encoder in the dotted box and keeping only the Cross-Modality Encoder.} 
\label{Fig.main2} 
\end{figure}

We perform the above three-stage pre-training on a simplified model of LXMERT (LXMERT-S). The simplified process of the LXMERT model is shown in Figure \ref{Fig.main2}. The Cross-Modality Encoder of LXMERT-S is identical to the LXMERT. We obtain the Simplified LXMERT (LXMERT-S) by removing the Object-Relationship Encoder and Language Encoder. The image features and text features are directly input to the Cross-Modality Encoder in the LXMERT-S.

By removing the single modal coding layer in LXMERT, the 12-layer LXMERT is simplified to a 5-layer LXMERT-S. The amounts of parameters in simplified LXMERT-S are only 45.9\% of the original model, and the whole experiment can be completed on a single GPU. The three-stage pre-training method is also fully applicable to other pre-training models.
\subsection{Stage 1: Word-based Pre-training}
\label{sect:pdf}
The first stage of pre-training focuses on learning the correspondence between text token units and image objects to help the model mine fine-grained information. To this end, we design the appropriate pre-training tasks and corresponding dataset for this phase of pre-training.

\paragraph{Pre-training Tasks}We design an Image Features Random Shuffle (IFRS) pre-training task to enhance the pre-training of the token layer, based on the existing Masked Language Modeling (MLM)~\citep{Chen:19}, Masked Region Feature Regression (MRFR)~\citep{Chen:19} and Masked Object Classification (MOC)~\citep{qi:20}.

Image Features Random Shuffle (IFRS): Given a set of image regions \(R\!=\!\{r_1, r_2, r_3…r_m\}\), which are obtained by adding a fully-connected (FC) layer to the regions of interest (ROIs) and projecting them to the hidden size, a feature triplet is three consecutive features in R, e.g.  \(t_j\!=\!(r_i, r_{i+1}, r_{i+2}) \). A shuffle on a triplet is to randomly change the order of features in the triplet with a probability of 5\%. For example, the triplet \(t_j\) is shuffled as \(t_j^{[S]}\!=\!(r_{i+1}, r_{i+2}, r_i)\!=\!(r_i^{[S]}, r_{i+1}^{[S]}, r_{i+2}^{[S]})\).The shuffled triplet \(t_j^{[S]}\) is used as input for the network, and the corresponding output is converted to the dimensionality of 
ROIs to obtain \({h_\theta}{(t_j^{[S]})}\!=\!( {h_\theta}{(r_i^{[S]})}, {h_\theta}{(r_{i+1}^{[S]})}, {h_\theta}{(r_{i+2}^{[S]})})\). The ROIs extracted by Faster-RCNN corresponding to the original \(t_j\) is \({f_\theta} {(t_j)}\!=\!( {f_\theta} {(r_i) },{f_\theta} {(r_{i+1}} ) ,{f_\theta} (r_{i+2}))\),We use the L2 loss to calculate the distance between the network output \({h_\theta}(t_j^{[S]})\) and \({f_\theta} (t_j)\) as in the following equation.
\begin{equation}
\tiny{{L}={E_{(W,R){\sim}D}}{{\sum_{k=0}^{k=K}}{\sum_{i=k'}^{i={k'}+2}}||{h_\theta}{(r_i^{[S]} )}\!-\!{f_\theta} {(r_i)}{||}_2^2}}
\end{equation}
Where K is the number of shuffled triples.

Other pre-training tasks: We add the existing MLM, MRFR and MOC tasks to the token-based pre-training. MLM masks the token-level category labels of objects with a certain probability P, and the model predicts the masked category label based on the corresponding object feature on the image side. MRFR masks the object features, and the model predicts the original object-level features based on the text-side category label and information around the object. MOC predicts the category and attribute labels of the masked object features.
\paragraph{Training Data}We extract training data for IFRS task from caption-image pairs directly. For each image, 36 object features and their corresponding 36 category labels are provided by Faster-RCNN. These category labels have been unified with the text vocabulary, so they are all included in the text vocabulary. During training, the image side inputs the image features in sequence, and the text side inputs the category labels in the corresponding order. In the IFRS task, when the image side is shuffled, the order of the text side remains unchanged.

\subsection{Stage 2: Phrase-based Pre-training}
The previous stage explores the correspondence between the image objects and their category. This stage mines the correspondence between the image object and the phrase describing of the object. Since the phrase description usually contains richer information about the attributes of the object, such as "green old car", building a pre-training task based on the correspondence between the phrase and the object allows the model to obtain rich information about the attributes.
\paragraph{ Pre-training Tasks} We define a Topic of Image and Text for Phrase (TITP) pre-training task that more directly supports phrase-based information mining.

Topic of Image and Text for Phrase (TITP): Given a token sequence of image phrase-level description \(W\!=\!\{w_1, w_2, w_3…w_n\}\), object feature sequence \(R\!=\!\{r_1, r_2, r_3…r_m\}\), and correspondent  category label sequence \(L\!=\!\{l_1, l_2, l_3…l_m\}\) extracted by Faster-RCNN. Let topic set is \(topic\!=\!W{\cap}L\!=\!\{p_1, p_2…p_q\}\), and label set \(Y\!=\!\{y_1, y_2…y_v\}\), 
where \(v\) is the size of the vocabulary. If \(y_i{\in} topic\), then \(y_i\) is 1, otherwise \(y_i\) is 0. We add a FC layer to the multimodal representation to get \({s_\theta}(W,R)\),  predict the correct topic from the vocabulary size \(v\) categories, and use BCELoss to calculate the gap between the model output \({s_\theta}(W,R)\) and the label Y.
\begin{equation}
\tiny{{L}\!=\!{E_{(W\!,\!R){\sim}D}}{[1/v{\sum_{i=0}^{v-1}}({y_i}log{s_\theta}{(W\!,\!R)}\!+\!(1\!-\!{y_i})log(1\!-\!{s_\theta}(W\!,\!R))}}
\end{equation}

Other pre-training tasks: We add MLM, MRFR and MOC tasks to the phrase-based pre-training. MLM masks the attribute or category information of the phrase with a certain probability P, and the model predicts the masked information based on the corresponding object features. MRFR masks the object features of the image, and the model predicts the original object based on the phrase-level description on the text side and the surrounding object information, and MOC predicts the category and attribute of the object being masked based on the surrounding image features and the phrase-level description on the text side.
\paragraph{Training Data:}We obtain the corresponding training data based on the Visual Genome (VG)~\citep{Krishna:17} dataset, which contains a large number of phrases. We eliminate the phrases containing verbs. The remaining phrases are concatenated with commas to obtain a phrase-level description of the image. During training, the spliced VG phrase is used as input on the text side and 36 object features extracted by Faster-RCNN are input on the image side.
\subsection{Stage 3: Sentence-based Pre-training}
On the basis of the above token and phrase training, this stage uses the overall sentence-image correspondence relationship for pre-training to mine larger unit text-image related information.
\paragraph{Pre-training Tasks}we design two sentence-level pre-training tasks, Image-Text Matching Based on Hard Sample (ITM\_HS) and Topic of Image and Text for Sentence (TITS) described as follows.

Image-Text Matching Based on Hard Sample (ITM\_HS): The purpose of this task is to reduce the noise brought to the model when the text-image pair does not match. We retrieve the top M most similar images for each image from difficult samples file\footnote{The difficult sample comes from the difficult sample file in ViLBERT's Image-Text Retrieval task.} as the hard sample set. In the ITM\_HS task, each image is replaced with a randomly selected hard sample with probability of 50\% if the hard sample sets is not empty. If the set of current sample is empty, an image in the training set is randomly selected. Let the token sequence \(W\!=\!\{w_1, w_2, w_3…w_n\}\) and the image feature sequence \(R=\{r_1, r_2, r_3…r_m\}\), the label \(y{\in}\{0,1\}\) indicates whether the input image-text pair matches each other. We apply the FC layer on top of the multimodal representation to get \({s_\theta}(T,R)\), which is the matching score of the image and text.
\begin{equation}
\scriptsize{L\!=\!{E_{(W\!,\!R){\sim}D}}[ylog{s_\theta}(W\!,\!R)\!+\!(1\!-\!y)log(1\!-\!{s_\theta} (W\!,\!R))]}
\end{equation}

Topic of Image and Text for Sentence (TITS): The purpose of this task is to jointly predict the content described by both image and sentence information. Given a token sequence \(W\!=\!\{w_1, w_2, w_3…w_n\}\), an image feature sequence \(R\!=\!\{r_1, r_2, r_3…r_m\}\), category labels for object features \(L=\{l_1, l_2, l_3…l_m\}\), \(topic\!=\!W{\cap}L\!=\!\{p_1, p_2…p_q\}\), and label \(Y\!=\!\{y_1, y_2…y_v\}\), where \(v\) is the size of the vocabulary. If \(y_i{\in}topic,\) then \(y_i\) is 1, otherwise \(y_i\) is 0. We apply the FC layer on top of the multimodal representation, convert its dimension to the vocabulary size \(v\) to get \({s_\theta} (W,R)\), and use BCELoss to calculate the gap between the model output \({s_\theta} (W,R)\) and the label Y.
\begin{equation}
\tiny{L\!=\!{E_{(W\!,\!R){\sim}D}}[1/v{\sum_{k=0}^{k=K}}({y_i} log{s_\theta}(W\!,\!R)\!+\!(1\!-\! {y_i})log(1\!-\!{s_\theta}(W\!,\!R)))}
\end{equation}

Other pre-training tasks: We add the existing MLM, MRFR and MOC tasks to the sentence-based pre-training. MLM masks the information in the sentence and the model predicts the masked information based on the all information on the image side. MRFR masks the object features of the image and the model predicts the original object based on the overall information at the sentence level on the text side and the surrounding object information. MOC predicts the category and attribute of the masked object based on the image features and the text-side sentence-level description.
\begin{table*}
\centering
\begin{tabular}{m{2cm} m{2cm} m{3cm} m{3cm} m{3cm}}
\hline 
\multirow{2}{*}{model} & \multirow{2}{*}{parameter} & \multicolumn{2}{c}{training data} & \multirow{2}{*}{resource consumption} \\
\cline{3-4}
 &  & text-image pairs & text corpus  &    \\
\hline
VL-BERT & 134.8M & 3.3M & Wikipedia BooksCorpus  & {16 V100 GPUs} \\
Unified VLP & - & 3.3M & - & {8 V100 GPUs} \\
ViLBERT & 218.9M & 3.3M & - & {8 TitanX GPUs} \\
LXMERT & 183.5M & 9.18M &- & { 4 TitanX GPUs} \\
VisualBERT & 85.05M & 1.28M &- & - \\
ours & 84.3M & 1.08M & - & {1 TitanX GPUs} \\
\hline
\end{tabular}
\caption{Comparison of parameter size, training data and resource consumption between the model in this paper and some pre-trained models.}
\label{table1}
\end{table*}
\paragraph{Training Data}In this stage, the image and its corresponding caption are directly used as input, the sentence level information caption is input on the text side, and the 36 object features provided by Faster-RCNN are input on the image side.
\section{Experiments}
\label{sec:length}
\subsection{Pre-training Dataset}
In this paper, the model is pre-trained using the COCO dataset and part of the VG dataset, and only 1.08M text-image pairs are used, where 0.12M image-text pairs are used in token-based pre-training stage, 0.34M image-text pairs are used in phrase-based pre-training stage, and 0.62M image-text pairs are used in the sentence-based pre-training stage. All datasets we used are also used in initial LXMERT. Table \ref{table1} gives a comparison of the pre-training data, model parameters\footnote{We exclude the parameters of the word embedding and pre-training task and only count the number of parameters in the Transform part.} and computational resources with other models.
\subsection{Downstream Tasks and Data Sets}
Visual Question Answering (VQA): There are multiple datasets for VQA. We use three common used datasets: VQA V2.0~\citep{Goyal:17}, GQA~\citep{Hudson:19}, and NLVR2~\citep{Suhr:19}. Accuracy is used as to measure model performance.

Cross-modal Retrieval task: We choose Flickr30K~\citep{Young:14} dataset as the retrieval task data, and evaluate the performance of the model in Image Retrieval (IR), Text Retrieval (TR), Zero Shot Image Retrieval (ZS-IR), and Zero Shot Text Retrieval (ZS-TR) respectively, and the performance metric is the matching score of text and image pairs. Zero shot is to evaluate the performance of the pre-trained model directly on the test set without fine-tuning, and is used to evaluate the effect of the pre-trained model. Therefore ZS-IR and ZS-TR are directly loaded with model parameters to perform IR and TR tasks without fine-tuning.

In the fine-tuning stage, the multimodal representation of the model is passed through a FC layer as a joint representation of image and text to solve downstream tasks. For VQA tasks, we linearize the multimodal representation into the answer category dimension through the FC layer to predict the answer of each question. For the Image-Text Retrieval~\citep{Young:14} task, we randomly replace the image or text, construct three negative examples for an image-text pair, including two random negative examples and a hard sample, and use BCELoss to calculate the difference between the matching score and the text-image matching label .
\begin{table*}
\centering
\begin{tabular}{lllllll}
\hline
\multirow{2}{*}{model} & \multicolumn{2}{c}{VQA V2.0} & \multicolumn{2}{c}{GQA} & \multicolumn{2}{c}{NLVR2}   \\
 & test-dev     & test-std      & test-dev  &test-std    & val    & test-p       \\
 \hline
Unified VLP  & 70.5    & 70.7       & -         & -   & -  &-    \\
ViLBERT      & 70.55   & 70.92   & -    & -   & -   &-         \\
VisualBERT   & 70.8  & 71      & -   &-        & 67.4     & 67          \\
VL-BERT    & 71.16      & -      & -   &-      & -    & -            \\
\hline
LXMERT    & \textbf{72.42}       & \textbf{72.54}        & \textbf{59.8}  & \textbf{60.33}    & \textbf{74.9}   & 74.5         \\
\hline
ours    & 71.1\scriptsize{(98.18\%)}  & 71.18\scriptsize{(98.13\%)}  & 58.7\scriptsize{(98.16\%)} & 59.12\scriptsize{(97.99\%)} & 74.03\scriptsize{(98.84\%)} & \textbf{74.72}\scriptsize{(\(\uparrow0.22\%\))} \\
\hline
\end{tabular}
\caption{LXMERT-S results on VQA V2.0, GQA and NLVR2.}
\label{table2}
\end{table*}
\begin{table*}
\begin{tabular}{m{1.3cm}m{0.7cm}m{0.7cm}m{0.7cm}m{0.65cm}m{0.65cm}m{0.7cm}m{1.0cm}m{0.9cm}m{1.0cm}m{0.65cm}m{0.65cm}m{0.65cm}}
\hline
\multirow{2}{*}{model} & \multicolumn{3}{c}{IR(zero-shot)} & \multicolumn{3}{c}{TR(zero-shot)} & \multicolumn{3}{c}{IR} & \multicolumn{3}{c}{TR} \\
& R@1 & R@5 & R@10 & R@1 & R@5 & R@10 & R@1 & R@5 & R@10  & R@1 & R@5 & R@10  \\
\hline
ViLBERT & 31.86   & 61.12 & 72.8 & - & - & - & \textbf{58.2} & \textbf{84.9}  & \textbf{91.52} & -  & - & - \\
LXMERT  & 24 & 47.38 & 58.22 & 23.6  & 51.5  & 61.3      & -& -  & -  & -  & -  & -     \\
\hline
ours & \textbf{42.42} & \textbf{68.7}  & \textbf{77.92} & \textbf{49}  & \textbf{75}  & \textbf{81.8}  & 57.9\tiny{(99.4\%)}  & 83\tiny{(97.8\%)} & 88.7\tiny{(97.0\%)} & \textbf{64.6}   & \textbf{87.5}  & \textbf{90.4} \\
\hline
\end{tabular}
\caption{LXMERT-S results on Image-Text Retrieval task.}
\label{table3}
\end{table*}
\subsection{Baselines}
We compare our model with both single-stream multimodal pre-training models including Unified VLP~\citep{zhou:20}, VisualBERT~\citep{li:19} and VL-BERT~\citep{su:20} and two-stream models including ViLBERT~\citep{Batra:19} and LXMERT~\citep{Tan:19}. 
\paragraph{Unified VLP} Unified VLP uses a 12 layers of shared multi-layer transformer network for both encoding and decoding, which differs from many existing methods where the encoder and decoder are implemented using separate models. It conducts pre-training on the Conceptual Captions(CC)~\citep{Sharma:18} which has around 3.3 million image-text pairs, and requires 150 hours of training on the 8x V100 GPUS. Unified VLP includes only the MLM task when processing the comprehension tasks.
\paragraph{VisualBERT} VisualBERT contains 12 layers of transformer with 85.05M parameters. It first pre-trains on COCO Caption~\citep{lin:14} with MLM and ITM tasks and then continues pre-training on the target dataset with MLM task. The pre-training data sizes for VisuaBERT on the VQA V2.0 task are shown in Table \ref{table1}. For different downstream tasks, the second stage of pre-training needs to be re-trained.
\paragraph{VL-BERT} VL-BERT contains 12 layers of transformer with 134.8M parameters. It pre-trains on both visual-linguistic and text-only datasets. Samples are randomly drawn from both CC and BooksCorpus~\citep{kun15} \& English Wikipedia (at a ratio of 1:1) in each mini-batch. VL-BERT considers ITM to be harmful to downstream tasks and therefore only includes MLM and MOC tasks.
\paragraph{ViLBERT} ViLBERT extends the popular BERT architecture to a multi-modal two-stream model, processing both visual and textual inputs in separate streams that interact through co-attentional transformer layers. It trains on CC with MLM, MOC and ITM tasks.
\paragraph{LXMERT} LXMERT has a large-scale Transformer model that consists of three encoders and a large-scale pre-training data, including MS COCO, Visual Genome, VQA v2.0, GQA and VG-QA~\citep{zhu:16}. The pre-training requires 8.5 days on the 4x TitanX  GPUS. It also has many pre-training tasks, including MLM, MRFR, MOC, ITM and Image Question Answering (QA)~\citep{Tan:19}, and has achieved good results in downstream tasks, especially VQA tasks.

\subsection{Implementation Details}Our Transformer backbone is the same as LXMERT, where each Transformer block has 768 hidden units and 12 attention heads. Image features are extracted by Faster-RCNN~\citep{ren:15} model (with ResNet-101~\citep{Zhang:16} backbone) trained on Visual Genome (VG).

During pre-training, our model is trained for about 95 hours on 1 TitanX GPU, and takes Adam~\citep{Kingma:15} as the optimizer with a learning rate of 1e-5. We train the token-based model for 10 epochs with a batch size of 64, phrase-based model for 20 epochs with a batch size of 128 and sentence-based model for 20 epochs with a batch size of 128.

During Fine-tuning, the learning rate of all downstream tasks is 5e-5, and the batch size is 32. We fine-tune 6 epochs for VQA V2.0, 5 epochs for GQA, and 8 epochs for NLVR2 and Image-Text Retrieval tasks.

For hard samples in ITM\_HS task, we retrieve the top 100 most similar images from difficult samples file. For the masking strategies, we randomly mask 15\% tokens, 15\% object features. The codes of our models are available at \url{https://github.com/lttsmn/LXMERT-S}.

\subsection{Experimental Results}Table \ref{table2} gives the results of the model on the three VQA datasets, and Table \ref{table3} gives the results of the model on the Flickr30K Image-Text Retrieval dataset.

It can be seen from both Table \ref{table2} and \ref{table3} that the pre-training model proposed in this paper has achieved comparable performances with the existing large models under the condition of less training data, fewer parameters and less computing resource occupation. In some cases, our small model even outperforms the big one. For example, NLVR2 task is 0.22 higher than LXMERT on Test-P, and ZS-IR is 18.42 higher than LXMERT in R@1 under the premise that the model parameters are reduced by 54.1\% and the training data set is reduced by 88.24\%.
\begin{table*}
\centering
\begin{tabular}{m{1.1cm} m{1.5cm} m{3.2cm} m{0.9cm} m{0.9cm} m{0.9cm} m{0.8cm}  m{0.8cm} m{0.8cm} m{0.9cm}}
 \hline
 Stage(s) count & Stage(s) used & Tasks used& \small{VQA}  test\_dev & \small{GQA} test\_dev & \small{NLVR2} test-p & \small{IR} avg & \small{ZS-IR} avg & \small{TR} avg& \small{ZS-TR} avg\\
 \hline
 \small{None} &vanilla & None &68.1 &55.71&51.07&55.27&-&58.07&- \\
 \hline
 \multirow{4}{*}{\small{Single}}& \multirow{3}{*}{\small{S}} &\small{MLM MRFR MOC TITS ITM\_HS} & 70.25 & 57.66  &  70.23 & 73.86 & 54.64 & 78.3 & 59.73  \\
 \cline{3-10}
 & &- \small{ITM\_HS}& 69.87 & 57.48& 70.98 & 71.46 & 51.79 & 75.6 & 53.33\\
 \cline{3-10}
 & &- \small{ITM\_HS} \hphantom{x} - \small{TITS} & 69.79 & 57.47 & 70.73 & 70.17 & 49.38 & 74.1 & 50.3\\
 \cline{2-10}
& \small{T + P + S} & \small{MLM MRFR MOC ITM} & 70.1 & 57.58 & 72.24 &74.31 &59.31 &77.87 &63.33\\
 \hline
 \multirow{4}{*}{\small{Two}} &\multirow{2}{*}{\small{T \(\to\) S}}& \small{MLM MRFR MOC TITS ITM\_HS IFRS}  & 70.71 & 58.39 & 73.85 & 75.68 & 61.53 & 80.27 & 65.47 \\
 \cline{3-10}
 & &- \small{IFRS}&70.54 & 58.4 & 73.93 & 76.08 & 60.5 & 80.6& 65.03 \\
 \cline{2-10}
 & \multirow{2}{*}{\small{P \(\to\) S}}& \small{ MLM MRFR MOC TITS ITM\_HS TITP}& 70.58 & 57.96 &72.96 &74.81 &57.66 &79.5&61.13\\
 \cline{3-10}
 & &- \small{TITP}& 70.52 & 58.17& 71.18& 75.49 & 59.28 & 80.4 & 62.73 \\
\hline
\multirow{4}{*}{\small{Three}}&\multirow{2}{*}{\small{T \(\to\) P \(\to\) S}} &\small{MLM MRFR MOC TITS ITM\_HS IFRS TITP} & \textbf{71.1} &\textbf{58.7} & \textbf{74.72} & 76.55 &63.01 & 80.83 & \textbf{68.6}  \\
\cline{3-10}
 & &- \small{TITP} &71.01 & 58.3 & 74.48 & 76.07 & \textbf{63.07} & 80.96 & 67.77 \\
\cline{2-10}
& \small{S \(\to\) P \(\to\) T} & \small{MLM MRFR MOC TITS ITM\_HS IFRS TITP}& 69.43 & 57.98 &56.75 &71.03 & - &74.87 &-\\
\cline{2-10}
& \small{P \(\to\)  T \(\to\) S} & \small{MLM MRFR MOC TITS ITM\_HS IFRS TITP} & 70.92 & 58.05 & 73.62   &\textbf{76.69} &61.29 & \textbf{81.63}&67\\
\hline
\end{tabular}
\caption{\label{table4}
Use VQA, GQA, NLVR2, Image-Text Retrieval (Flickr30k) downstream tasks to evaluate the MSP method and pre-training tasks. Image-Text Retrieval uses the average value of R@1, R@5, R@10. }
\end{table*}
\subsection{Ablation Study}
Table \ref{table4} gives results of LXMERT-S on different tasks with different pre-training setting. The first column gives the number of stage(s) in pre-training. The second column gives the stage(s) used, where S for sentence stage, P for phrase stage, and T for token stage, \(T{\to}S\) means there are two stages including token-based pre-training first and then sentence-based pre-training. \(T{\to}P{\to}S\) means there are three stages including token-based pre-training first and then phrase-based pre-training and sentence-based pre-training last. T+P+S means to train all stages together. The third column gives the pre-training tasks used in the pre-training. We first give all the pre-training tasks used in the training stages used, then verify the validity of the pre-training tasks by removing a task based on all the pre-training tasks, “-” indicates that a pre-training task is removed.

From Table \ref{table4}, we can find: (1) With the orderly increase of the training phase, the performance of the model on downstream tasks is gradually improving; (2) The training granularity from small to large is the most effective training sequence; (3) The pre-training tasks we propose for each stage of pre-training can improve the performance of the model on downstream tasks, such as TITP improves VQA performance by 0.09, GQA performance by 0.4, NLVR2 performance by 0.24, IR performance by 0.48, and ZS-TR by 0.83. 

\section{Qualitative Analysis}
\begin{figure}
\centering 
\includegraphics[width=0.48\textwidth]{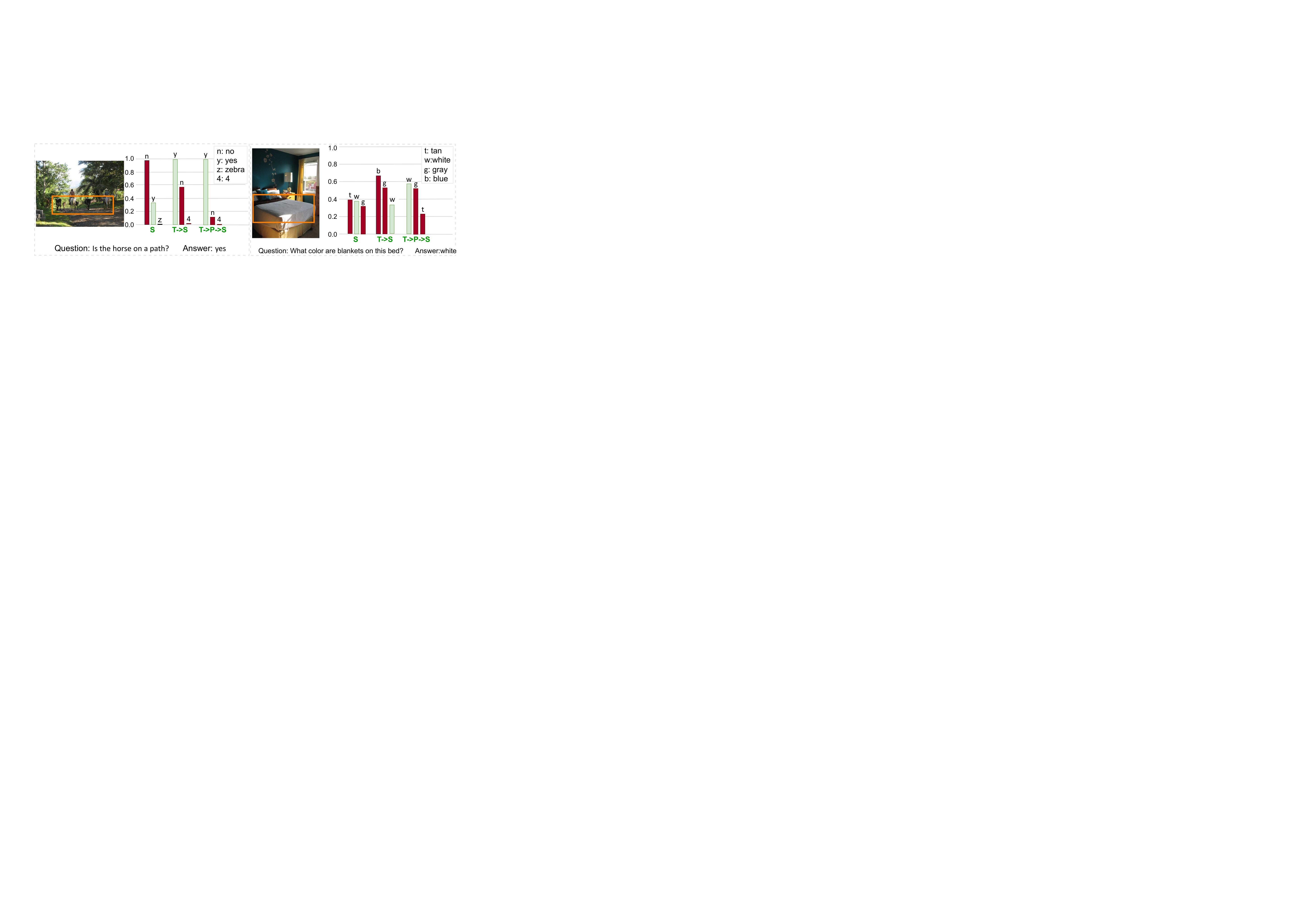}
\caption{VQA examples. We show the distribution of answers with top 3 scores at different pre-training stages.} 
\label{Fig.main6} 
\end{figure}

We visualize the impact of different pre-training stages on VQA and Image-Text Retrieval task by showing the answers probability distribution. For each example in Figure \ref{Fig.main6}, the left side is the input image of the model, and the right side is the probability distribution of the top3 scoring answers in different pre-training stages. 

\begin{figure}
\centering 
\includegraphics[width=0.48\textwidth]{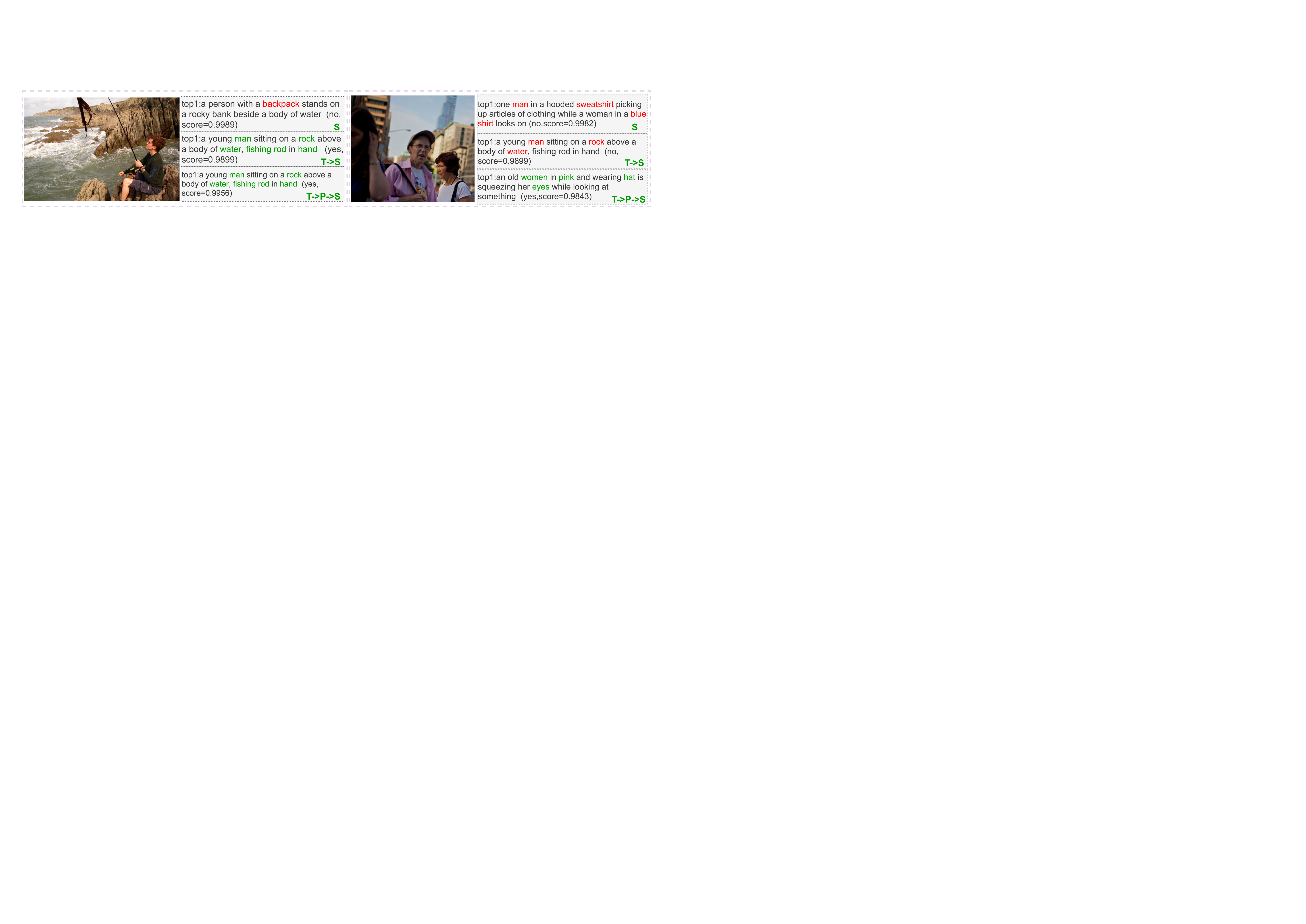}
\caption{Image-Text Retrieval examples. We show the distribution of caption with top 1 score at different pre-training stages.} 
\label{Fig.main7} 
\end{figure}

For Image-text Retrieval task, we select the top 1 caption for visualization. For each sample in Figure \ref{Fig.main7}, the left side is the input image and the right side is the highest scoring caption predicted by the model.

From both Figure \ref{Fig.main6} and \ref{Fig.main7}, we can find: (1) Token-based pre-training (S vs \(T{\to}S\)) helps the model to learn object information in the images. For example, in the left sample in Figure \ref{Fig.main6} and \ref{Fig.main7}, the model improves its performance on downstream tasks by adding token-based pre-training that makes the model focus on object information such as horses, man and rocks in the images; (2) Phrase-based pre-training (\(T{\to}S\) vs \(T{\to}P{\to}S\)) helps the model to learn information about the attributes of the objects. As shown in right-hand image in Figure \ref{Fig.main6} and \ref{Fig.main7}, the model pays attention to attribute information, i.e. blanket is white, clothes are pink, etc.

\section{Conclusion}In this paper, inspired by the idea of curriculum learning, we propose a MSP method, which uses information at different granularities from word, phrase to sentence in both texts and images to pre-train a model in stages, we also design pre-training tasks suitable for each stage of pre-training, IFRS task for word-based pre-training, TITP task for phrase-based pretraining, and TITS task for sentence-based pretraining. Experimental results on several VQA datasets as well as one cross-modal retrieval dataset show that our method achieves similar or even better performance than a larger model in terms of accuracy in all downstream tasks under the premise that the model parameters are reduced by 54.1\% and the training data set is reduced by 88.24\%. In future work, we will add the above training method to other simplified pre-trained models to further explore the effectiveness of MSP method.
\section*{Acknowledgments}

We would like to thank anonymous reviewers for their suggestions and comments. The work was supported by the National Natural Science Foundation of China (NSFC62076032), the Cooperation Poject with Beijing SanKuai Technology Co., Ltd and the National Key Research and Development Program of China (2020YFF0305302). We would like to thank Dr. Huixing Jiang and his colleagues.

\bibliographystyle{IEEEtran}

\begin{thebibliography}{32}
\expandafter\ifx\csname natexlab\endcsname\relax\def\natexlab#1{#1}\fi

\bibitem[{Alberti et~al.(2019)Alberti, Ling, Collins, and Reitter}]{alberti:19}
Chris Alberti, Jeffrey Ling, Michael Collins, and David Reitter. 2019.
\newblock \href {https://doi.org/10.18653/v1/D19-1219} {Fusion of detected
  objects in text for visual question answering}.
\newblock In \emph{{EMNLP-IJCNLP} 2019}, pages 2131--2140.

\bibitem[{Bengio et~al.(2009)Bengio, Louradour, Collobert, and
  Weston}]{bengio:09}
Yoshua Bengio, J{\'e}r{\^o}me Louradour, Ronan Collobert, and Jason Weston.
  2009.
\newblock Curriculum learning.
\newblock In \emph{Proceedings of the 26th annual international conference on
  machine learning}, pages 41--48.

\bibitem[{Brown et~al.(2020)Brown, Mann, Ryder, Subbiah, Kaplan, Dhariwal,
  Neelakantan, Shyam, Sastry, Askell et~al.}]{brown:20}
Tom~B Brown, Benjamin Mann, Nick Ryder, Melanie Subbiah, Jared Kaplan, Prafulla
  Dhariwal, Arvind Neelakantan, Pranav Shyam, Girish Sastry, Amanda Askell,
  et~al. 2020.
\newblock Language models are few-shot learners.
\newblock \emph{arXiv preprint arXiv:2005.14165}.

\bibitem[{Chen et~al.(2019)Chen, Li, Yu, Kholy, Ahmed, Gan, Cheng, and
  Liu}]{Chen:19}
Yen{-}Chun Chen, Linjie Li, Licheng Yu, Ahmed~El Kholy, Faisal Ahmed, Zhe Gan,
  Yu~Cheng, and Jingjing Liu. 2019.
\newblock \href {http://arxiv.org/abs/1909.11740} {{UNITER:} learning universal
  image-text representations}.
\newblock \emph{CoRR}, abs/1909.11740.

\bibitem[{Devlin et~al.(2019)Devlin, Chang, Lee, and Toutanova}]{Devlin:18}
Jacob Devlin, Ming{-}Wei Chang, Kenton Lee, and Kristina Toutanova. 2019.
\newblock \href {https://doi.org/10.18653/v1/n19-1423} {{BERT:} pre-training of
  deep bidirectional transformers for language understanding}.
\newblock In \emph{{NAACL-HLT} 2019}, pages 4171--4186.

\bibitem[{Goyal et~al.(2017)Goyal, Khot, Summers{-}Stay, Batra, and
  Parikh}]{Goyal:17}
Yash Goyal, Tejas Khot, Douglas Summers{-}Stay, Dhruv Batra, and Devi Parikh.
  2017.
\newblock \href {https://doi.org/10.1109/CVPR.2017.670} {Making the {V} in
  {VQA} matter: Elevating the role of image understanding in visual question
  answering}.
\newblock In \emph{{CVPR} 2017}, pages 6325--6334. {IEEE} Computer Society.

\bibitem[{He et~al.(2016)He, Zhang, Ren, and Sun}]{Zhang:16}
Kaiming He, Xiangyu Zhang, Shaoqing Ren, and Jian Sun. 2016.
\newblock \href {https://doi.org/10.1109/CVPR.2016.90} {Deep residual learning
  for image recognition}.
\newblock In \emph{{CVPR} 2016}, pages 770--778. {IEEE} Computer Society.

\bibitem[{Hudson and Manning(2019)}]{Hudson:19}
Drew~A. Hudson and Christopher~D. Manning. 2019.
\newblock \href {https://doi.org/10.1109/CVPR.2019.00686} {{GQA:} {A} new
  dataset for real-world visual reasoning and compositional question
  answering}.
\newblock In \emph{{CVPR}2019,}, pages 6700--6709.

\bibitem[{Jiao et~al.(2020)Jiao, Yin, Shang, Jiang, Chen, Li, Wang, and
  Liu}]{Jiao:19}
Xiaoqi Jiao, Yichun Yin, Lifeng Shang, Xin Jiang, Xiao Chen, Linlin Li, Fang
  Wang, and Qun Liu. 2020.
\newblock \href {https://doi.org/10.18653/v1/2020.findings-emnlp.372}
  {Tinybert: Distilling {BERT} for natural language understanding}.
\newblock In \emph{{EMNLP} 2020}, pages 4163--4174.

\bibitem[{Kingma and Ba(2015)}]{Kingma:15}
Diederik~P. Kingma and Jimmy Ba. 2015.
\newblock Adam: {A} method for stochastic optimization.
\newblock In \emph{{ICLR} 2015}.

\bibitem[{Krishna et~al.(2017)Krishna, Zhu, Groth, Johnson, Hata, Kravitz,
  Chen, Kalantidis, Li, Shamma, Bernstein, and Fei{-}Fei}]{Krishna:17}
Ranjay Krishna, Yuke Zhu, Oliver Groth, Justin Johnson, Kenji Hata, Joshua
  Kravitz, Stephanie Chen, Yannis Kalantidis, Li{-}Jia Li, David~A. Shamma,
  Michael~S. Bernstein, and Li~Fei{-}Fei. 2017.
\newblock Visual genome: Connecting language and vision using crowdsourced
  dense image annotations.
\newblock \emph{Int. J. Comput. Vis.}, 123(1):32--73.

\bibitem[{Lan et~al.(2020)Lan, Chen, Goodman, Gimpel, Sharma, and
  Soricut}]{Chi:20}
Zhenzhong Lan, Mingda Chen, Sebastian Goodman, Kevin Gimpel, Piyush Sharma, and
  Radu Soricut. 2020.
\newblock \href {https://openreview.net/forum?id=H1eA7AEtvS} {{ALBERT:} {A}
  lite {BERT} for self-supervised learning of language representations}.
\newblock In \emph{{ICLR} 2020}.

\bibitem[{Li et~al.(2019)Li, Yatskar, Yin, Hsieh, and Chang}]{li:19}
Liunian~Harold Li, Mark Yatskar, Da~Yin, Cho{-}Jui Hsieh, and Kai{-}Wei Chang.
  2019.
\newblock \href {http://arxiv.org/abs/1908.03557} {Visualbert: {A} simple and
  performant baseline for vision and language}.
\newblock \emph{CoRR}, abs/1908.03557.

\bibitem[{Li et~al.(2020)Li, Yin, Li, Zhang, Hu, Zhang, Wang, Hu, Dong, Wei,
  Choi, and Gao}]{li:20}
Xiujun Li, Xi~Yin, Chunyuan Li, Pengchuan Zhang, Xiaowei Hu, Lei Zhang, Lijuan
  Wang, Houdong Hu, Li~Dong, Furu Wei, Yejin Choi, and Jianfeng Gao. 2020.
\newblock \href {https://doi.org/10.1007/978-3-030-58577-8\_8} {Oscar:
  Object-semantics aligned pre-training for vision-language tasks}.
\newblock In \emph{{ECCV} 2020}, volume 12375, pages 121--137.

\bibitem[{Lin et~al.(2014)Lin, Maire, Belongie, Hays, Perona, Ramanan,
  Doll{\'{a}}r, and Zitnick}]{lin:14}
Tsung{-}Yi Lin, Michael Maire, Serge~J. Belongie, James Hays, Pietro Perona,
  Deva Ramanan, Piotr Doll{\'{a}}r, and C.~Lawrence Zitnick. 2014.
\newblock Microsoft {COCO:} common objects in context.
\newblock In \emph{{ECCV} 2014}, volume 8693, pages 740--755. Springer.

\bibitem[{Lu et~al.(2019)Lu, Batra, Parikh, and Lee}]{Batra:19}
Jiasen Lu, Dhruv Batra, Devi Parikh, and Stefan Lee. 2019.
\newblock Vilbert: Pretraining task-agnostic visiolinguistic representations
  for vision-and-language tasks.
\newblock In \emph{{NeurIPS} 2019}, pages 13--23.

\bibitem[{Michel et~al.(2019)Michel, Levy, and Neubig}]{Michel:19}
Paul Michel, Omer Levy, and Graham Neubig. 2019.
\newblock \href
  {http://papers.nips.cc/paper/9551-are-sixteen-heads-really-better-than-one}
  {Are sixteen heads really better than one?}
\newblock In \emph{NeurIPS 2019}, pages 14014--14024.

\bibitem[{Ordonez et~al.(2011)Ordonez, Kulkarni, and Berg}]{Ordonez:11}
Vicente Ordonez, Girish Kulkarni, and Tamara~L. Berg. 2011.
\newblock Im2text: Describing images using 1 million captioned photographs.
\newblock In \emph{{NIPS} 2011}, pages 1143--1151.

\bibitem[{Qi et~al.(2020)Qi, Su, Song, Cui, Bharti, and Sacheti}]{qi:20}
Di~Qi, Lin Su, Jia Song, Edward Cui, Taroon Bharti, and Arun Sacheti. 2020.
\newblock \href {http://arxiv.org/abs/2001.07966} {Imagebert: Cross-modal
  pre-training with large-scale weak-supervised image-text data}.
\newblock \emph{CoRR}, abs/2001.07966.

\bibitem[{Radford et~al.(2018)Radford, Narasimhan, Salimans, and
  Sutskever}]{radford:18}
Alec Radford, Karthik Narasimhan, Tim Salimans, and Ilya Sutskever. 2018.
\newblock Improving language understanding by generative pre-training (2018).

\bibitem[{Radford et~al.(2019)Radford, Wu, Child, Luan, Amodei, and
  Sutskever}]{radford:19}
Alec Radford, Jeffrey Wu, Rewon Child, David Luan, Dario Amodei, and Ilya
  Sutskever. 2019.
\newblock Language models are unsupervised multitask learners.
\newblock \emph{OpenAI blog}, 1(8):9.

\bibitem[{Ren et~al.(2015)Ren, He, Girshick, and Sun}]{ren:15}
Shaoqing Ren, Kaiming He, Ross~B. Girshick, and Jian Sun. 2015.
\newblock Faster {R-CNN:} towards real-time object detection with region
  proposal networks.
\newblock In \emph{{NIPS} 2015}, pages 91--99.

\bibitem[{Sharma et~al.(2018)Sharma, Ding, Goodman, and Soricut}]{Sharma:18}
Piyush Sharma, Nan Ding, Sebastian Goodman, and Radu Soricut. 2018.
\newblock \href {https://doi.org/10.18653/v1/P18-1238} {Conceptual captions:
  {A} cleaned, hypernymed, image alt-text dataset for automatic image
  captioning}.
\newblock In \emph{{ACL} 2018}, pages 2556--2565. Association for Computational
  Linguistics.

\bibitem[{Su et~al.(2020)Su, Zhu, Cao, Li, Lu, Wei, and Dai}]{su:20}
Weijie Su, Xizhou Zhu, Yue Cao, Bin Li, Lewei Lu, Furu Wei, and Jifeng Dai.
  2020.
\newblock \href {https://openreview.net/forum?id=SygXPaEYvH} {{VL-BERT:}
  pre-training of generic visual-linguistic representations}.
\newblock In \emph{{ICLR} 2020}.

\bibitem[{Suhr et~al.(2019)Suhr, Zhou, Zhang, Zhang, Bai, and Artzi}]{Suhr:19}
Alane Suhr, Stephanie Zhou, Ally Zhang, Iris Zhang, Huajun Bai, and Yoav Artzi.
  2019.
\newblock \href {https://doi.org/10.18653/v1/p19-1644} {A corpus for reasoning
  about natural language grounded in photographs}.
\newblock In \emph{{ACL} 2019,}, pages 6418--6428. Association for
  Computational Linguistics.

\bibitem[{Tan and Bansal(2019)}]{Tan:19}
Hao Tan and Mohit Bansal. 2019.
\newblock \href {https://doi.org/10.18653/v1/D19-1514} {{LXMERT:} learning
  cross-modality encoder representations from transformers}.
\newblock In \emph{{EMNLP-IJCNLP} 2019}, pages 5099--5110.

\bibitem[{Vaswani et~al.(2017)Vaswani, Shazeer, Parmar, Uszkoreit, Jones,
  Gomez, Kaiser, and Polosukhin}]{Vaswani:17}
Ashish Vaswani, Noam Shazeer, Niki Parmar, Jakob Uszkoreit, Llion Jones,
  Aidan~N. Gomez, Lukasz Kaiser, and Illia Polosukhin. 2017.
\newblock Attention is all you need.
\newblock In \emph{{NIPS} 2017}, pages 5998--6008.

\bibitem[{Young et~al.(2014)Young, Lai, Hodosh, and Hockenmaier}]{Young:14}
Peter Young, Alice Lai, Micah Hodosh, and Julia Hockenmaier. 2014.
\newblock \href
  {https://tacl2013.cs.columbia.edu/ojs/index.php/tacl/article/view/229} {From
  image descriptions to visual denotations: New similarity metrics for semantic
  inference over event descriptions}.
\newblock \emph{Trans. Assoc. Comput. Linguistics}, 2:67--78.

\bibitem[{Zafrir et~al.(2019)Zafrir, Boudoukh, Izsak, and
  Wasserblat}]{Zafrir:19}
Ofir Zafrir, Guy Boudoukh, Peter Izsak, and Moshe Wasserblat. 2019.
\newblock \href {http://arxiv.org/abs/1910.06188} {{Q8BERT:} quantized 8bit
  {BERT}}.
\newblock \emph{CoRR}, abs/1910.06188.

\bibitem[{Zhou et~al.(2020)Zhou, Palangi, Zhang, Hu, Corso, and Gao}]{zhou:20}
Luowei Zhou, Hamid Palangi, Lei Zhang, Houdong Hu, Jason~J. Corso, and Jianfeng
  Gao. 2020.
\newblock \href {https://aaai.org/ojs/index.php/AAAI/article/view/7005}
  {Unified vision-language pre-training for image captioning and {VQA}}.
\newblock In \emph{{AAAI} 2020}, pages 13041--13049. {AAAI} Press.

\bibitem[{Zhu et~al.(2016)Zhu, Groth, Bernstein, and Fei{-}Fei}]{zhu:16}
Yuke Zhu, Oliver Groth, Michael~S. Bernstein, and Li~Fei{-}Fei. 2016.
\newblock \href {https://doi.org/10.1109/CVPR.2016.540} {Visual7w: Grounded
  question answering in images}.
\newblock In \emph{{CVPR} 2016}, pages 4995--5004. {IEEE} Computer Society.

\bibitem[{Zhu et~al.(2015)Zhu, Kiros, Zemel, Salakhutdinov, Urtasun, Torralba,
  and Fidler}]{kun15}
Yukun Zhu, Ryan Kiros, Richard~S. Zemel, Ruslan Salakhutdinov, Raquel Urtasun,
  Antonio Torralba, and Sanja Fidler. 2015.
\newblock \href {http://arxiv.org/abs/1506.06724} {Aligning books and movies:
  Towards story-like visual explanations by watching movies and reading books}.
\newblock \emph{CoRR}, abs/1506.06724.

\end{thebibliography}

\end{document}